\documentclass{article}
\usepackage[utf8]{inputenc}
\usepackage{authblk}
\usepackage{setspace}
\usepackage[margin=1.25in]{geometry}
\usepackage{graphicx}
\graphicspath{ {./figures/} }
\usepackage{subcaption}
\usepackage{amsmath}
\usepackage[T1]{fontenc}
\usepackage{float}
\usepackage{amsfonts}
\usepackage{amssymb}
\usepackage[T1]{fontenc}
\usepackage{lmodern,tikz,lipsum}
\usepackage{fancyhdr}
\usepackage{changepage}
\usepackage{hyperref}
\usepackage{tocloft}
\usepackage{wallpaper}
\usepackage{lipsum} 
\usepackage{titlesec} 

\usepackage{algorithm}
\usepackage{algpseudocode}

\title{How to unlearn a learned Machine Learning model ?}

\author[1]{Seifeddine Achour}

\date{2024}

\begin{document}

\maketitle

\begin{abstract}
In contemporary times, machine learning (ML) has sparked a remarkable revolution across numerous domains, surpassing even the loftiest of human expectations. However, despite the astounding progress made by ML, the need to regulate its outputs and capabilities has become imperative. A viable approach to address this concern is by exerting control over the data used for its training, more precisely, by unlearning the model from undesired data. In this article, I will present an elegant algorithm for unlearning a machine learning model and visualize its abilities. Additionally, I will elucidate the underlying mathematical theory and establish specific metrics to evaluate both the unlearned model's performance on desired data and its level of ignorance regarding unwanted data.

\end{abstract}


\section{Introduction}
Similar to human learning, the current machine learning concept consists in examining the existing data, keep training over it in order to master it, which enables the model to make decisions concerning new unknown data. During the training process, the usual ML algorithm tries to minimize an error function with respect to model parameters, using as a reference the whole provided training data. Now, our problem is how to adjust the model parameters to keep only the desired samples' history. This will be realized by defining a specific objective function that involves minimizing the error on desired data and maximizing it on undesired data. Therefore the model not only will forget about the unwanted data but also should maintain the same performance on desired ones or even become more accurate as we will see later in the article.

\section{Modeling and approach}
The usual mathematical approach to training an ML model consists in suggesting a model that aligns the most with the data type and distribution (linear, polynomial, decision tree, neural network...)\cite{muk}, defining a relevant criterion (MSE, MAE, RMSE...)\cite{cort}, choosing an optimizer(SGD, Adam...)\cite{wang}, then look for the parameters of the model that minimize the most the error function (criterion). Through this process, the model will master the all training data and get ready to generate outputs for new input data. In order to readjust the model parameters to forget about the unpreferred data, I suggest another objective function which I named Ethical MSE (EMSE):

\[
\sum \limits _{i \in wanted}(y_i - \hat{y}_i)^2 - \sum \limits _{i \in unwanted} \log(1-\frac{1}{\sqrt{\pi}}\exp(-(y_i - \hat{y}_i)^2))
\]

Let's understand the mathematical theory behind this formula. Let $y_1,\ldots,y_n$ be the training data and $\mathbf{w}$ the model parameters. Assuming that $y_1,\ldots,y_n$ are independent and the probability of guessing the output of the new data sample $y_{n+1}$ knowing the training data $y_1,\ldots,y_n$ and the model parameters vector $\mathbf{w}$ is defined by  $L = P(y_{n+1} | y_1, \ldots, y_n, \mathbf{w}) = \prod \limits_{i=1}^{n} P(y_{n+1} | y_i, \mathbf{w}) = \prod \limits _{i \in wanted} P(y_{n+1} | y_i, \mathbf{w}) \prod \limits _{i \in unwanted} P(y_{n+1} | y_i, \mathbf{w})$.\\ 
Assuming that $P(y_{n+1} | y_i, \mathbf{w}) = \frac{1}{\sqrt{2\pi}\sigma} \exp\left(-\frac{(y_i - \hat{y}_i)^2}{2\sigma^2}\right)$ the Gaussian distribution $N(y_i,\sigma^2)$, we have to maximize the likelihood of wanted data and minimize the likelihood of the unwanted data. This means looking for: 
\[
\arg\max{\prod \limits _{i \in wanted} P(y_{n+1} | y_i, \mathbf{w}) \prod \limits _{i \in unwanted} (1-P(y_{n+1} | y_i, \mathbf{w}))}
\]
=
\[
\arg\min -\log({\prod \limits _{i \in wanted} \frac{1}{\sqrt{2\pi}\sigma} \exp\left(-\frac{(y_i - \hat{y}_i)^2}{2\sigma^2}\right) \prod \limits _{i \in unwanted} (1-\frac{1}{\sqrt{2\pi}\sigma} \exp\left(-\frac{(y_i - \hat{y}_i)^2}{2\sigma^2}\right))})
\]
=
\[
\arg\min \sum \limits _{i \in wanted}\frac{(y_i - \hat{y}_i)^2}{2\sigma^2} - \sum \limits _{i \in unwanted} \log(1-\frac{1}{\sqrt{2\pi}\sigma}\exp(-\frac{(y_i - \hat{y}_i)^2}{2\sigma^2})
\]\\
For $\sigma = \frac{1}{\sqrt{2}}$\\
   =

\[
\arg \min \sum \limits _{i \in wanted}(y_i - \hat{y}_i)^2 - \sum \limits _{i \in unwanted} \log(1-\frac{1}{\sqrt{\pi}}\exp(-(y_i - \hat{y}_i)^2))
\]\\

So this criterion allows the trained model to maintain the maximum of its knowledge about desired data and forget the maximum of the unwanted data as we will see in the next section.

\section{Experiment and Results}
In this section, I will describe the conducted experiment that demonstrates the practical implementation of the EMSE (Ethical MSE) criterion. We will explore a real case and analyze the significant outcomes that arise from applying this criterion.

In the experiment, I trained a polynomial model on data which is divided into two parts: desired data and unwanted data. In the beginning, the model is trained on both wanted and unwanted data. After training is finished, the algorithm of unlearning, which uses the EMSE, will be applied. 

\subsection{Training on all data}
Let's first examine the result of training on all the data:

\begin{figure}[H]
    \centering
    \includegraphics[width=13.5cm,height=10cm]{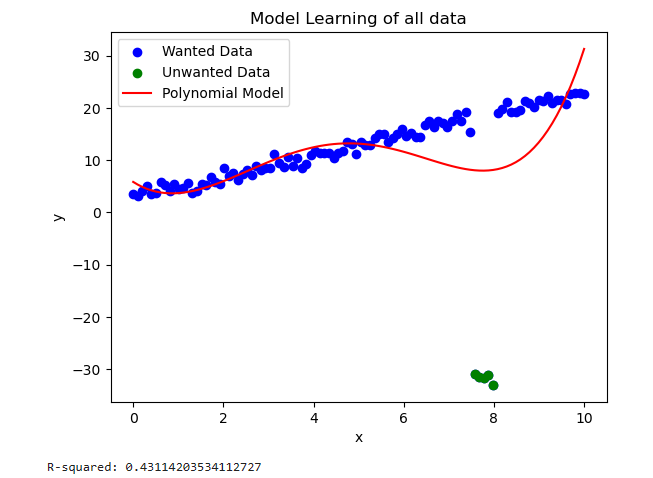}
    \caption{Model trained on all data }
    \label{fig:example}
\end{figure}

We see in the plot that the model mastered somehow the wanted data but in a certain region it gets closer to the unwanted data which means that effectively, it is aware of it and trained on it. We see also that being aware of the unwanted data reduced its performance on the desired data where $R-squared = 0.43$.

\subsection{Machine Unlearning}
So far we have the parameters of the model which was trained on all data including the unwanted one. Now, our purpose is to unlearn the pre-trained model from the unwanted data using the EMSE criterion which will maximize the error of the model on the undesired data and minimize it on the desired one as we see in the following figure:

\begin{figure}[H]
    \centering
    \includegraphics[width=13.5cm,height=10cm]{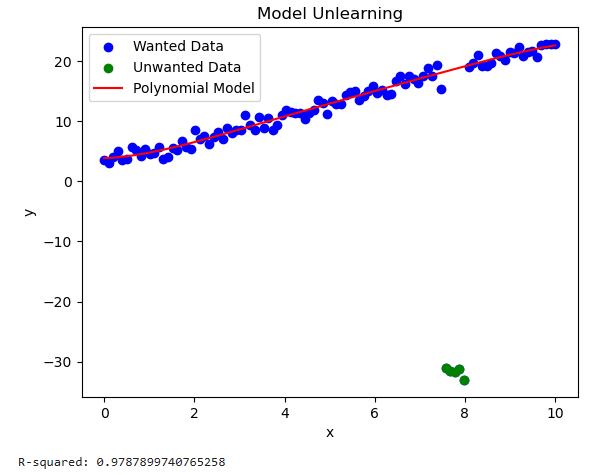}
    \caption{Model trained on all data }
    \label{fig:example}
\end{figure}

The plot shows clearly how the model didn't show any inclination toward the green data which means he forgot totally about it after the unlearning process. This process has provided another benefit, as we see that the R-squared of the wanted data passed from $0.43$ to $0.98$ which proves notably the power of the Unlearning algorithm.

\subsection{Huge amount of unwanted data}
In this part, the algorithm will be tested in the case where the amount of unwanted data is significantly comparable to the wanted data size.

\begin{figure}[H]
  \centering
  \begin{subfigure}{0.498\textwidth}
    \includegraphics[width=\linewidth]{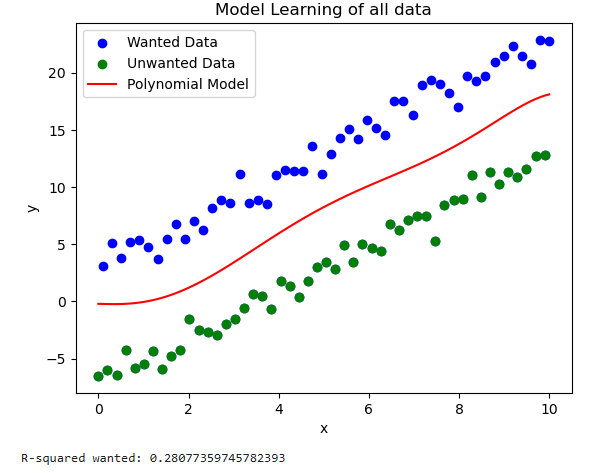}
    \caption{Learned Model}
    \label{fig:plot1}
  \end{subfigure}
  \hfill
  \begin{subfigure}{0.49\textwidth}
    \includegraphics[width=\linewidth]{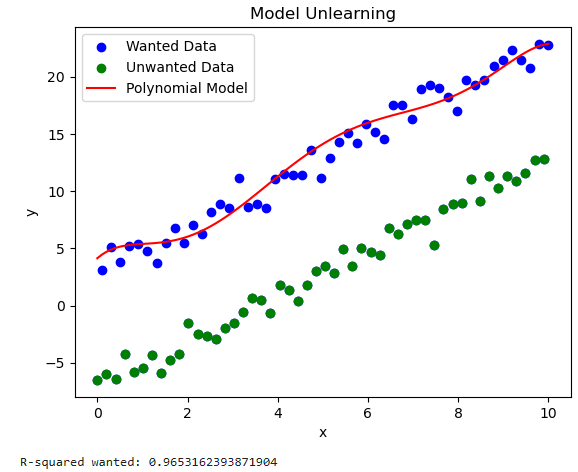}
    \caption{Unlearned Model}
    \label{fig:plot2}
  \end{subfigure}

  \caption{Model curve before and after unlearning}
  \label{fig:4plots}
\end{figure}

According to the previous plots, the unlearning algorithm showed its robustness even with a huge amount of unwanted data. Also, again the performance of the model on the wanted data increased significantly where the R-squared passed from 0.28 to 0.96.

\section{Metrics and Discussion}

In the previous section, we saw the high efficiency of the algorithm that uses the EMSE criterion, where it succeed very well in erasing the existence of the unwanted data from the model function. Also, it increased the model performance on wanted data, because by forgetting the unwanted data the model could focus more on the wanted one. Otherwise in order to measure the effective efficiency of the new unlearned model, we should define a metric that indicates its degree of excellence in mastering the wanted data and mediocrity with unwanted one.

\subsection{The Exponential R-squared} 
One of the metrics used to evaluate a model's performance is the R-squared metric. It measures the proportion of the total variation in the dependent variable explained by the regression model \cite{camer}. This metric is good to determine the representativeness of the regression model.
In purpose to confirm to which point the model forgot about the undesired data points, I defined a metric called Exponential R-squared which presents the probability that the model is unrepresentative of the unwanted data

\[
\text{Exponential }R^2 =1 - \exp(-(1-R^2))
\]
Where 
\[
R^2 = 1 - \frac{{\text{RSS}}}{{\text{TSS}}}
\]
And
\[ 
\text{RSS(Residual Sum of Squares)} = \sum \limits _{i \in unwanted} (y_i - \hat{y}_i)^2
\]
And
\[
\text{TSS(Total Sum of Squares)} = \sum \limits _{i \in unwanted}  (y_i - \bar{y})^2
\]\\
When the Exponential $R^2$ value is closer to 1, it indicates that the model is less representative, implying that it has neglected or failed to capture the irrelevant data.

\subsection{Fair R-squared}
So far we have an unlearned model and a metric that can determine the probability of the unrepresentativeness of a model to certain data. Now what we want is a global metric that tells us the representativeness of the wanted data and the unrepresentativeness of unwanted data at the same time. here comes the role of the Fair R-squared metric which is the probability that the model is both representative for the wanted data and unrepresentative for the unwanted data at the same time. Its formula is as follows:
\[
\text{Fair R-squared} = P(\text{representative } \cap \text{ unrepresentative}) 
\]\\
If we assume that the two events are independent then:
\[
\text{Fair R-squared} = P(\text{representative }) P(\text{ unrepresentative}) = (1 - \text{Exponential }R^2_{wanted})\text{ Exponential }R^2_{unwanted} 
\]\\

\subsection{Limits and discussion}
The Fair R-squared metric defined previously will contribute directly to our understanding of the model performance in mastering the desired data and in ignoring the other data. Otherwise, its formula is based on the assumption of independence between both previous tasks. This can not always be satisfied as shown in the following figure:

\begin{figure}[H]
    \centering
\includegraphics[width=13.5cm,height=10cm]{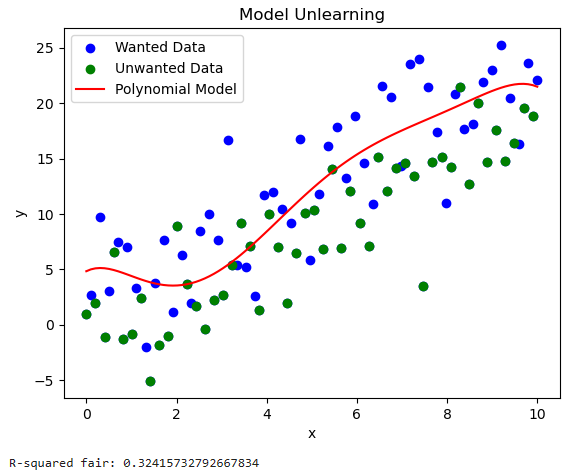}
    \caption{Dependence between wanted and unwanted data }
    \label{fig:example}
\end{figure}

We can notice from the previous plot that distinguishing between the two types of data would be very challenging, which means that mastering blue data points and ignoring green data points aren't independent at all, so we should check the independence assumption. Additionally, the model can't easily avoid unwanted data while mastering the other one. Increasing the model complexity would give it more freedom to adjust itself but still can't increase a lot the Fair R-squared. Also, the choice of the right standard deviation $\sigma$ could be a potential research direction to investigate its impact on the model.

\section{Conclusion}
In this article, I presented a novel algorithm capable of unlearning specific data from a trained model, employing the newly introduced criterion EMSE. Furthermore, I introduced a metric called Fair R-squared, which serves as an evaluation tool for gauging the model's proficiency in comprehending the relevant data while disregarding the irrelevant data. Lastly, I emphasized the significance of verifying the assumption of data independence to appropriately employ the appropriate formula for Fair R-squared and gain a proper understanding of the inherent limitations in data segregation.


\end{document}